\documentclass{article} 
\usepackage{iclr2020_conference,times}

\usepackage{amsmath,amsfonts,bm}

\def\eqref#1{equation~\ref{#1}}

\def\1{\bm{1}}

\DeclareMathAlphabet{\mathsfit}{\encodingdefault}{\sfdefault}{m}{sl}
\SetMathAlphabet{\mathsfit}{bold}{\encodingdefault}{\sfdefault}{bx}{n}

\usepackage{hyperref}
\usepackage{url}
\usepackage{multirow}
\usepackage[normalem]{ulem}
\useunder{\uline}{\ul}{}
\usepackage{graphicx}
\newcommand{\para}[1]{{\vspace{5pt} \bf \noindent #1 \hspace{2pt}}}

\usepackage{subfig}

\title{MSD: Multi-Self-Distillation Learning via Multi-classifiers within
	Deep Neural Networks}

\author{Yunteng Luan, Hanyu Zhao, Zhi Yang \& Yafei Dai 
	\\
Department of Electronics Engineering and Computer Science\\
Peking University\\
Beijing, China \\
\texttt{\{luanyunteng,zhaohanyu,yangzhi,dyf\}@pku.edu.cn} \\
}

\begin{document}

\maketitle

\begin{abstract}
As the development of neural networks, more and more deep neural networks are adopted in various tasks, such as image classification.
However, as the huge computational overhead, these networks could not be applied on mobile devices or other low latency scenes.
To address this dilemma, multi-classifier convolutional network is proposed to allow faster inference via early classifiers with the corresponding classifiers.
These networks utilize sophisticated designing to increase the early classifier accuracy.
However, n\"{a}ively training the multi-classifier network could hurt the performance (accuracy) of deep neural
networks as early classifiers throughout interfere with the
feature generation process.

In this paper, we propose a general training framework
named multi-self-distillation learning (MSD), which mining knowledge of different classifiers within the same network and increase every classifier accuracy.
Our approach can be applied not only to multi-classifier networks,
but also modern CNNs (e.g., ResNet Series) augmented with additional side branch classifiers.
We use \emph{sampling-based branch augmentation} technique to transform a single-classifier network into a multi-classifier network.
This reduces the gap of capacity between different classifiers, and improves the effectiveness of applying MSD.
Our experiments show that MSD improves the accuracy of various networks:
enhancing the accuracy of every classifier significantly for existing multi-classifier network (MSDNet), 
improving vanilla single-classifier networks with internal classifiers with high accuracy,
while also improving the final accuracy.
\end{abstract}

\section{Introduction}
Deep convolutional
networks (CNNs) are already adopted in
a diverse set of visual recognition tasks such as image classification \cite{huang2018gpipe,krizhevsky2012imagenet,tan2019efficientnet}. 
With the ever-increasing demand for improved performance, the development of deeper networks has greatly increased the latency and computational cost of inference.
These costs prevents models from being deployed on resource
constrained platforms (e.g., mobile phones) or applications requiring a short response time (e.g., self-driving cars).
To lessen these increasing costs, multi-classifier network architectures \cite{larsson2016fractalnet, teerapittayanon2016branchynet} are proposed to lessen the inference costs by allowing prediction
to quit the network early when samples can already be inferred with high
confidence.
Multi-classifier networks posit
that the easy examples do not require the full power and
complexity of a massive DNN.
So rather than attempting to
approximate existing networks with weights pruning and quantization, they introduce multiple early classifiers
throughout a network, which are applied on the features of the particular layer they are attached
to.

However, the introduction of early classifiers into network could interfere negatively with later classifiers~\cite{huang2017multi}.
How to overcome this drawback is the key to design multi-classifier network architectures. For example, use dense connectivity to connects each layer with all subsequent layers.
However, we make an observation that the later classifiers may not always be able to rightly classify test examples correctly predicted by earlier ones.
For example, about 25.4\% test samples predicted correctly by the first classifier of MSDNets~\cite{huang2017multi} cannot rightly predicted by any later classifiers (including the final classifier) on on CIFAR100 dataset. This implies that increasing learning independence in multi-classifier network also hinders the knowledge transfer among multiple classifiers.

To solve this dilemma, we propose a novel \emph{multi-self-distillation learning} framework
where classifiers in a multi-classifier network learn collaboratively
and teach each other throughout the training process. One significant advantage of multi-self-distillation learning framework
is that it doesn't need other collaborative student models required in traditional mutual learning \cite{zhang2018deep}.
All the classifiers within the network itself are trained as student
models who effectively pools their
collective estimate of the next most likely classes with different levels of features. Specifically, each classifier is trained with
three losses: a conventional supervised learning loss, a
prediction mimicry loss that aligns each classifier’s class posterior with
the class probabilities of other classifiers, and a feature mimicry loss that induces all the classifiers' feature maps to fit the feature
maps of the deepest classifier. 
The last loss consides heterogeneous cohorts
consisting of mixed deepest classifier and shallow classifier, and enables the learning more efficiently with (more or less) bias towards the prowerful (deepest) classifier. 

MSD learning helps each classifier
to obtain more discriminating features, which enhances
the performance of other classifiers in return.
With such learning, the model not only requires
less training time but also can
accomplish much higher accuracy, as compared with other
learning methods (such as traditional knowledge distillation and mutual learning).
In general, this framework can also be applied to improve the performance of single-classifier CNNs by adding additional early-classifier branches at certain locations throughout the original network. 
For simplicity,
in this paper we focus on typical group-wise networks, such as Inception and ResNet Series, where CNN architectures are assembled as the stack of basic block structures.
Each of group shares
similar structure but with different weights and filter numbers, learning features of fine
scale in early groups and coarse scale in later groups (through repeated convolution, pooling, and
strided convolution).

With such kind of group-wise network architecture, we propose a \emph{sampling-based branch augmentation method} to address the design considerations of (1) the locations of early-classifier branches, and (2) the structure of a early-classifier branch as well as its size and depth.
Specifically, we add early-classifier branches after different group to allow the samples to quit after processing a subset of groups.
We determine the structure of a specific early-classifier branch by performing intra-and-inter-group sampling over the remaining network deeper than the attached point of the branch. The basic idea of this sampling-based method is to enable the network path quits from any early-classifier branch to approximate that classifier from the main branch, i.e., the baseline (original) network. This reinforces the efficiency and learning capabilities of individual branch classifiers.
Also, our method provides a single neural network
quits at different depth, permitting dynamic inference specific to test examples.

Extensive experiments are carried out on two image-classification datasets. 
The results show that, for specially designed network with multiple classifiers, the MSD learning improves the performance of every classifier by a large margin with the same network architecture.
Further, by argument modern convolutional neural networks with early-classifier branches, the MSD learning significantly improves the performance of these network at no expense
of response time. 3.2\% accuracy increment is obtained on
average for ResNet Series, varying from 1.47\% in ResNeXt as minimum
to 4.56\% in ResNet101 as maximum. Finally, compared with
self distillation by the deepest classifier~\cite{Zhang_2019_ICCV}, collaborative MSD learning by all classifiers achieves better performance. 


In summary, the main contributions of this paper are:
\begin{itemize}
	\item We propose a MSD learning framework which provides a simple but effective
	way to improve the performance of a network with multiple classifiers.
	\item We provide an classifier-branch augmentation method to permit modern CNNs to be optimized with the proposed MSD learning.
	\item We conduct experiments for different kinds of CNNs and training methods on the task of image classification to prove
	the generalization of this learning method.
\end{itemize}
\section{Related Work}

\subsection{Knowledge Distillation}
KD (knowledge distillation) is a model compression technique proposed by \cite{bucilu2006model}. 
And it was utilized for neural networks in \cite{hinton2015distilling}. 
Traditional KD try to transfer a big pretrained teacher network's knowledge to a smaller student network.
In details, it compute a KL loss between the teacher and student output distributions. 
And this loss provides additional regularisation and supervision for the student.
In this case, the student accuracy may be higher than the teacher.
Various KD techniques have been proposed. 
FitNet\cite{romero2014fitnets} propose a hint loss to minimize the distance of feature maps between teacher and network,
and then it uses classical KD technique to train the re-initial student network.
AT\cite{zagoruyko2016paying} explores FitNet using two kinds of attention techniques.
\cite{lopes2017data} proposes a KD solution in case of unable to obtain training data.
\cite{yim2017gift} defines an FSP matrix to represent knowledge and proposes a approach to transfer.
\cite{mirzadeh2019improved} discusses the gap between teacher and student in KD, 
and proposes a cascade KD technique.
\cite{Zhang_2019_ICCV} proposes self-distillation, and this method does not need a pretrained teacher.
Our work is possibly most closely related to this work, however, self-distillation focus on improving the 
final accuracy, and it only use the final classifier to teach the middle classifiers. 
While our approach aims to improve each classifier accuracy and use multiple teachers.
DML\cite{zhang2018deep} also does need a pretrained teacher. It trains multiple networks at the same time, 
and make them teach each other.
However, this method introduces more training burden, 
and the small network must waiting for multiple large networks.

\subsection{Multi-classifier Networks}
Various prior studies explore ACT (adaptive computation time) networks. Recently, a new branch of ACT is multi-classifier network.
Multi-classifier network is first proposed by BranchyNet \cite{teerapittayanon2016branchynet}.
It is a network equipped with multiple early classifier connected with a backbone. 
As Figure \ref{archi} illustrates, it has three early classifiers. This kind of architecture has many advantages. 
On the one hand, it provide multiple tiny-networks to satisfy different capacity and latency needs 
without hurting the final classifier accuracy.
On the other hand, it can be treated as an ensemble network. And because these classifiers share the same backbone network,
multi-classifier network is more efficient computation than traditional ensemble network.
FractalNet \cite{larsson2016fractalnet} proposes a multi-path network, and each path consumes different computation, achieve different accuracy.
A permutation of these paths provide various latency and performance.
SkipNet \cite{wang2018skipnet} proposes a adaptive network architecture based on ResNet, 
and it skips unnecessary ResNet blocks utilizing reinforcement learning.
MSDNet \cite{huang2017multi} propose a novel multi-classifier network inspired by DenseNet, 
and it adopts multi-scale technique to increase early classifier accuracy.
\section{Method}
In this section, we give an example to illustrate how to apply \emph{sampling-based branch augmentation} to a non-multi-classifier network.
And then we give a detailed description of our proposed multi-self-distillation learning technique based on our example.

\subsection{Sampling-based branch augmentation}
\begin{figure*}[h]
	\centering
	\includegraphics[scale=0.4]{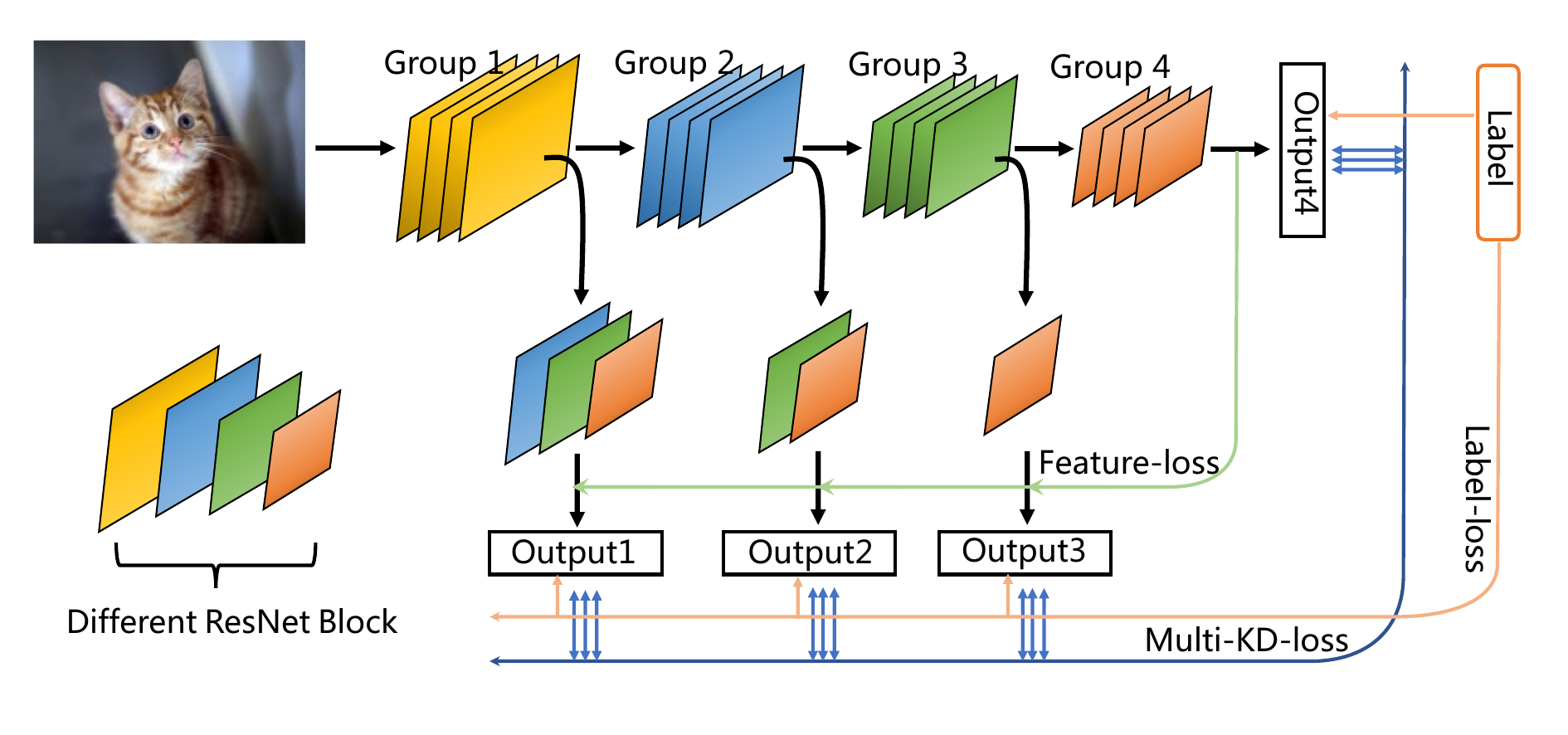}
	\caption{A ResNet-style network equipped with multiple classifiers. 
		We produce a early classifier behind to each layer block. 
		Every block has multiple ResNet layers, and will shrink feature map width and height dimension, 
		increase channel dimension.
		In order to make the early-classifier's feature map dimension changes more smoothly,  
		we equip the first, second and third classifier with 3, 2 and 1 ResNet block, respectively.}
	\label{archi}
\end{figure*}
In Figure \ref{archi}, we illustrate a modified ResNet-style network, which is equipped with multiple classifiers.
In Resnet-style network, each layer group contains multiple ResNet blocks, 
and each layer group resizes the prior feature map dimension: shrinks feature map width and height dimension, 
increases channel dimension, in details.
In order to make the early-classifier's feature map dimension changing pattern is similar with the backbone network,
we equip the first, second and third classifier with 3, 2 and 1 ResNet layer, respectively.
And these extra ResNet layers is a instance of our proposed \emph{sampling-based branch augmentation} architecture.
The amount of computation added by the \emph{sampling-based branch augmentation} is negligible relative to the entire network.
However, these blocks bring a huge increase in accuracy, according to the experiment results.

\subsection{Multi-Self-Distillation Learning}

\para{Formulation.}
We assume a dataset $X = \{x_i\}$ with $M$ classes $Y=\{y_i\}, y_i \in \{1, 2, ..., M\}$, 
and a network with $N$ classifiers. For the $n$ classifier, its output is $a^n$. 
We use softmax to compute the predicted probability $p$:
\begin{equation}
p_{i}^{n} = \frac{exp(a_{i}^{n})} {\sum{exp(a_{j}^{n}})}
\end{equation}
where $p_{i}^{n}$ represents the $i_{th}$ class probability of the $n$ classifier.

\para{Loss Function.} MSD loss consist of three parts, \textit{label loss}, \textit{kd loss} and \textit{feature loss}.

\textit{Label loss.} The first loss comes from the label $y$ provided by the dataset.
For each classifier, we compute cross entropy between $p^{n}$ and $y$. 
In this way, the label $y$ directs each classifier's proper probability as high as possible.
As there are multiple classifier, we sum each cross entropy loss:
\begin{equation}
loss_1 = \sum_{n=1}^{N}CrossEntropy(p^n, y)
\end{equation}

\textit{KD loss.} In classical knowledge distillation\cite{hinton2015distilling}, there is a student network $Net^s$ with an output $a^s$,
and a teacher network $Net^t$ with an output $a^t$. The KD loss for $Net^s$ is computed by:
\begin{equation}
loss_{KD} = KL(p_{\tau}^{s}, p_{\tau}^{t})
\end{equation}
where KL is Kullback-Leibler divergence, and $p_{\tau}^{s}$ and $p_{\tau}^{t}$ are soften probabilities:
\begin{equation}
p_{\tau, i}^{s} = \frac{exp(a_{i}^{s}/\tau)} {\sum{exp(a_{j}^{s}/\tau})},
p_{\tau, i}^{t} = \frac{exp(a_{i}^{t}/\tau)} {\sum{exp(a_{j}^{t}/\tau})}
\end{equation}
where $\tau$ represents temperature. 
A higher temperature gives softer probability distribution and more knowledge to the student network.

For each classifier, we treat all the other $N-1$ classifier as its teacher networks.
As different teacher provide different knowledge, we could achieve a more robust and accurate network.
We use the average losses as each classifier KD loss:
\begin{equation}
loss_2 = \frac{1}{N-1} \cdot \sum_{i=1}^{N}\sum_{j\neq i}^{N}KL(q_{\tau}^{i}, q_{\tau}^{j})
\end{equation}
and $\tau$ depends on the class number $M$.

\textit{Feature loss.} Inspired by Fitnets\cite{romero2014fitnets}, we compute the L2 distance between the feature maps before the final FC layer.
On one hand, the hint loss also provide knowledge for the early classifiers, and helps convergence. 
On the other hand, as \cite{mirzadeh2019improved} says, when the student does not have the sufficient capacity or mechanics to mimic the teacher’s behavior, the knowledge distillation may be not efficient. And the hint loss forces student to approach the weight distribution of teachers, in other words, it reduce the gap between teacher and student.
\begin{equation}
loss_3 = \sum_{i=1}^{N-1}\left \| F_{i}-F_{N} \right \|_{2}^{2}
\end{equation}
where $F_i$ represents the feature maps before the FC layer.

\para{Training}
During training, we compute the sum of above three parts of loss. And to balance the three parts loss, 
we introduce two hyper-parameters $\alpha$ and $\beta$:
\begin{equation}
\begin{split}
totalloss &= (1 - \alpha) \cdot loss_1 + \alpha \cdot loss_2 + \beta \cdot loss_3
\\&=\left ( 1-\alpha  \right ) \cdot \sum_{n=1}^{N}CrossEntropy(p^n, y)
\\& + \alpha \cdot \frac{1}{N-1} \cdot \sum_{i=1}^{N}\sum_{j\neq i}^{N}KL(q_{\tau}^{i}, q_{\tau}^{j})
\\& + \beta \cdot \sum_{i=1}^{N-1}\left \| F_{i}-F_{N} \right \|_{2}^{2}
\end{split}
\end{equation}

As the feature loss is used to help the early classifiers convergence as the beginning, 
a big $\beta$ may hurt the network performance at the end of training. We adopt a cosine annealing policy for $\beta$:
\begin{equation}
\beta = 0.5 \cdot (1 + cos(epoch/total \cdot \pi) \cdot (\beta_{begin} - \beta_{end})) + \beta_{end}
\end{equation}
where $\beta_{begin}, \beta_{end}$ represents initial $\beta$ and final $\beta$.
Experiments show this policy is better than a constant $\beta$.
\section{Experiments}
In this section, we elaborate experiments on different networks and datasets to demonstrate our approach.
All experiment code is implemented by PyTorch. And we would release our code later.

\begin{table*}[h]
	\begin{center}
		\begin{tabular}{|c|c|c|c|c|c|c|}
			\hline
			Networks           & Method         & Classifier 1 & Classifier 2 & Classifier 3 & Classifier 4 & Classifier 5 \\ \hline
			\multirow{2}{*}{MSDNet-1} & joint-training & 62.40          & 65.37          & 69.74          & 71.88          & 73.63          \\ \cline{2-7} 
			& multi-self-dis & \textbf{64.13} & \textbf{67.86} & \textbf{71.35} & \textbf{73.65} & \textbf{74.93} \\ \hline
			\multirow{2}{*}{MSDNet-2} & joint-training & 64.44          & 69.28          & 71.88          & 73.38          & 74.75          \\ \cline{2-7} 
			& multi-self-dis & \textbf{66.63} & \textbf{70.79} & \textbf{73.30} & \textbf{73.99} & \textbf{75.09} \\ \hline
		\end{tabular}
	\end{center}
	\caption{Accuracy comparison on MSDNet\cite{huang2017multi} (CIFAR100).
		MSDNet-1 set base=1, step=1, block=5, mode=lin\_grow, and MSDNet-2 set base=3, step=3, block=5, mode=even.
		More network details are described in paper\cite{huang2017multi}.}
	\label{msdnet comparison}
\end{table*}

\begin{table}[h]
	\begin{center}
		\begin{tabular}{|c|c|c|c|c|c|c|c|}
			\hline
			Networks                  & Naive-Train                & Method         & Classifier1                              & Classifier2                              & Classifier3               & Classifier4        \\ \hline
			&                         & self-dis-orign & {\color[HTML]{FE0000} 67.85}                & {\color[HTML]{FE0000} 74.57}                & 78.23                        & 78.64                    \\ \cline{3-7} 
			\multirow{-2}{*}{ResNet18}       & \multirow{-2}{*}{77.09} & multi-self-dis & {\ul \textbf{78.93}}                        & {\ul \textbf{79.63}}                        & {\ul \textbf{80.13}}         & {\ul \textbf{80.26}}  \\ \hline
			&                         & self-dis-orign & {\color[HTML]{FE0000} 68.23}                & {\color[HTML]{FE0000} 74.21}                & {\color[HTML]{FE0000} 75.23} & 80.56                    \\ \cline{3-7} 
			\multirow{-2}{*}{ResNet50}       & \multirow{-2}{*}{77.68} & multi-self-dis & {\ul \textbf{78.6}}                         & {\ul \textbf{80.36}}                        & {\ul \textbf{81.67}}         & {\ul \textbf{81.78}} \\ \hline
			&                         & self-dis-orign & {\color[HTML]{FE0000} 69.45}                & {\color[HTML]{FE0000} 77.29}                & 81.17                        & 81.23                    \\ \cline{3-7} 
			\multirow{-2}{*}{ResNet101}      & \multirow{-2}{*}{77.98} & multi-self-dis & {\ul \textbf{78.29}}                        & {\ul \textbf{80.47}}                        & {\ul \textbf{82.75}}         & {\ul \textbf{82.54}} \\ \hline
			&                         & self-dis-orign & {\color[HTML]{FE0000} 68.84}                & {\color[HTML]{FE0000} 78.72}                & 81.43                        & 81.61                    \\ \cline{3-7} 
			\multirow{-2}{*}{ResNet152}      & \multirow{-2}{*}{79.21} & multi-self-dis & {\color[HTML]{FE0000} {\ul \textbf{77.1}}}  & {\ul \textbf{80.98}}                        & {\ul \textbf{82.83}}         & {\ul \textbf{82.74}} \\ \hline
			&                         & self-dis-orign & {\color[HTML]{FE0000} 68.85}                & {\color[HTML]{FE0000} 78.15}                & {\ul \textbf{80.98}}         & 80.92                    \\ \cline{3-7} 
			\multirow{-2}{*}{WRN20-8} & \multirow{-2}{*}{79.76} & multi-self-dis & {\color[HTML]{FE0000} {\ul \textbf{76.81}}} & {\color[HTML]{FE0000} {\ul \textbf{78.60}}} & 80.62                        & {\ul \textbf{81.23}} \\ \hline
			&                         & self-dis-orign & {\color[HTML]{FE0000} 72.54}                & 81.15                                       & 81.96                        & 82.09                    \\ \cline{3-7} 
			\multirow{-2}{*}{WRN44-8} & \multirow{-2}{*}{79.93} & multi-self-dis & {\color[HTML]{FE0000} {\ul \textbf{77.11}}} & {\ul \textbf{79.95}}                        & {\ul \textbf{82.17}}         & {\ul \textbf{82.28}} \\ \hline
		\end{tabular}
	\end{center}
	\caption{Accuracy comparison with self-distillation on CIFAR100 dataset. 
		Naive-Training represents training the network with only cross-entropy loss.
		Self-dis-orign represents self-distillation results on the original paper\cite{Zhang_2019_ICCV}.
		Multi-self-dis represents our approach results.}
	
	\label{comp}
\end{table}

\subsection{Dataset}
We conduct experiments on two popular datasets respectively. 
CIFAR100 contain 60 thousand RGB images of 32x32 pixels with 100 classes.
And 50 thousand images for training, 10 thousand images for test.
We use random cropping, random horizontal flipping and normalization for preprocessing.

\subsection{Multi-classifier Networks}

There are many works focus on designing multi-classifier network architecture. 
MSDNet proposes a multi-scale networks for resource efficient image classification and achieves SOTA results.
In this subsection, we select some kinds of MSDNet network to verify our approach's effects. 
Note that we do not change any training details such as lr, training epochs, etc. from the original paper.

From the Table \ref{msdnet comparison}, it is observed that our approach beats the original training on every classifier,
and achieves average over 1\% increment. 
This proves that MSD is effective on multi-classifier networks.

\subsection{Non-Multi-Classifier Network}
We evaluate our approach with CIFAR100 dataset on multiple classical and efficient Networks,
including ResNet18, ResNet50, ResNet101, ResNet152, and WideResNet20-8, WideResNet44-8.
We treat self-distillation as baseline as it achieves SOTA results on these models.

The experiment results is reported in Table \ref{comp}. 
Baseline is the original network accuracy by naive training method.
From the table, wo could summarize some conclusions.
1) All final classifiers (Classifier 4/4) based on our approach beat self-distillation and naive training, 
and achieve average nearly 1\% and 3.2\% increment, respectively.
1) All middle classifiers except one (Classifier3/4 of WRN20-8) beat self-distillation.
Especially, the first classifier achieve average 8.5\% increment.
3) The accuracy difference between the first classifier and the final classifier is very small, 
although the first classifier only takes a little part of FLOPs compared with the final classifier.

\section{Conclusion}

We proposed a novel training framework called MSD (Multi-self-distillation) to mine the inherent knowledge within the model to improve its accuracy.
We conducted various experiments on multi-classifier networks, single-classifier networks and different datasets,
to prove its advantages compared with vanilla, self-distillation techniques.
Moreover, MSD does not need too much extra training cost or other neural network helps,
compared with traditional knowledge transfer and knowledge distillation.
In order to apply MSD on single-classifier networks,
we also proposed \emph{sampling-based branch augmentation} technique to extend single-classifier to multi-classifier.
By this way, the original network not only achieves higher accuracy on the final classifier,
but also could be utilized as an effective multi-classifier network.

\clearpage

\bibliography{iclr2020_conference}
\bibliographystyle{iclr2020_conference}

\end{document}